# On the Liveliness of Artificial Life

*Yong Zher Koh and Maurice HT Ling*

## Abstract

There has been on-going philosophical debate on whether artificial life models, also known as digital organisms, are truly alive. The main difficulty appears to be finding an encompassing and definite definition of life. By examining similarities and differences in recent definitions of life, we define life as *"any system with a boundary to confine the system within a definite volume and protect the system from external effects, consisting of a program that is capable of improvisation, able to react and adapt to the environment, able to regenerate parts of itself or its entirety, with energy system comprises of non-interference sets of secluded reactions for self-sustenance, is considered alive or a living system. Any incomplete system containing a program and can be re-assembled into a living system; thereby, converting the re-assembled system for the purpose of the incomplete system, are also considered alive."* Using this definition, we argue that digital organisms may not be the boundary case of life even though some digital organisms are not considered alive; thereby, taking the view that some form of digital organisms can be considered alive. In addition, we present an experimental framework based on continuity of the overall system and potential discontinuity of elements within the system for testing future definitions of life.

## 1  Introduction

The definition of life has been one of the greatest philosophical questions of mankind. In recent years, this debate had intensified due to the discovery of naturally occurring biological entities, such as viruses and prions, which lie at the boundary of what we consider as living. "Are viruses alive?" has turned out to be the largest vote swinging debate in an introductory course to microbiology [1], with 79% of the students changing their opinions before and after the debate compared to genetic engineering (56% opinion swing) and childhood immunization (25% opinion swing). Using boundary cases as classification criteria, also known as the decision boundary or decision surface, had been well established [2]. The reason that "Are viruses alive?" makes an interesting debate is simple – we will have established certain definite criteria separating what is alive from what is not by answering this question. To put this question on a decision boundary is, in fact, asking, "are viruses the first case of living object or the last case of a non-living object?" Despite using tools of molecular biology to study the genetics of viruses [3], the debate continues. It is unlikely that a definitive conclusion can be reached in the near future.

In the computational world, we are faced with similar philosophical questions – Are computer viruses considered to be living? Are artificial life (AL) forms alive? Christopher Langton coined the term "artificial life" in 1986 as a discipline to study life and its processes as we understand, which may go beyond the creations of nature [4]. The general view taken by *alifers* (artificial life researchers) is that the carbon-based biological life on Earth is merely one manifestation or implementation of life out of the many other forms it may have taken. For example, if at the very early days of Earth, the very first macromolecule is silicon-based instead of carbon-based, we may be eating "silicohydrates" instead of carbohydrates today. Similarly, if life starts in Jupiter, the organisms may be drinking liquid hydrogen instead of dihydro-monoxide (which is water). Hence, carbon-based life forms are implementations of life in the current ecosystem that we are in. If we are in a different planet where there is only arsenic and no phosphorus, the resulting implementation of life will use arsenic instead, as a species of bacterium was found to do in order to survive in an arsenic-rich environment of Mono Lake in California, USA [5].

However, there has been a continuing debate in the field of AL of whether computer-simulated life forms,

also known as digital organisms, are considered alive [6-11]. The main difficulty appears to be finding an encompassing and definite definition of life. Machery [12] had proposed that the act of defining life might be pointless after all.

In this communication, we attempt to argue that some forms of digital organisms can be considered alive under the current framework of determining life and non-life. We propose that some forms of digital organisms can actually be considered as alive as bacteria are.

## 2 What is Life? And a New Definition of Life

The definition of life has long been of great interest to both biologists and philosophers alike [13]. Popa [14] and Barbieri [15] had presented a list of more than 100 such definitions (see Appendix A for a compiled list from Popa [14], Barbieri [15], and a collated list of definitions from the year 2003). In this section, we shall review recent developments in this area. This analysis is by no means exhaustive but presents a snap shot of the current state of philosophy.

Koshland [16], in an attempt to define life that encompasses extra-terrestrial life, lists seven "pillars of life". Firstly, there must be a program, which is an organized plan that describes both the ingredients themselves and the kinetics of the interactions among ingredients as the living system persists through time. Secondly, there must be a way to change or improvise this program. Thirdly, the living entity must be separable from the non-living spaces. This is also known as compartmentalization. Fourthly, there must be a source of energy to enable some form of metabolism. Fifthly, the entity must be able to regenerate or reproduce to compensate for the wear and tear on a living system. In this case, regeneration can mean self-repair and recovery. Sixthly, the entity must be able to adapt to environmental hazards. Lastly, each set of chemical process must be secluded from each other in some way to prevent interference.

Edward Trifonov had presented two different approaches to this issue. In the first instance [17], he examined the usage of amino acids based on thermodynamics and looked at early proteins. Using this method, he defined life as imprecise replication. In the next instance [18], he examined commonalities between 123 definitions from Popa [14] and Barbieri [15] and defined life as "self-reproduction with variations." It is interesting to note that both definitions are essentially the same thing as replication is synonymous with self-reproduction and imprecision inevitably results in variations. Moreover, these concepts, replication and variations, are the same concepts of regeneration and improvisation respectively, as proposed by Koshland [16]. Hence, it can be said that Trifonov's definitions are proper subsets of Koshland's definition.

There are several other recent definitions that are similar to Trifonov [18], which render themselves assimilated under Koshland's definition. Ruiz-Mirazo et al. [19] defined a living being as any autonomous system with open-ended evolutionary capacities. Evolutionary capabilities can be considered as improvisation whereas autonomous can be considered as compartmentalization, metabolism or seclusion under Koshland's definition. James Carroll [20] suggests that the simplest form of life exists in the form of self-amplifying, autocatalytic reactions, which can be considered as secluded or compartmentalized, self-regenerating, metabolic reactions under Koshland's definition. Lazcano [21] defines life as a self-sustaining chemical system (such as, one that turns environmental resources into its own building blocks) that is capable of undergoing natural selection, which can be considered as Carroll's simplest life form undergoing improvisation. Hence, Lazcano's definition is also a proper subset of under Koshland's definition.

Zhuravlev and Avetisov [22] examined terrestrial life to abstract key components and events leading to the emergence of life. These include a living state resulting from the interaction between matter and energy carriers, a living system resulting from self-reproduction, and a living process to interact and adapt to the environment. A living state, a living system, and a living process in Zhuravlev and Avetisov's definition corresponds to metabolism, regeneration, adaptability of compartmentalized systems in Koshland's definition respectively.

Macklem and Seely [23] defines life as a "self-contained, self-regulating, self-organizing, self-reproducing, interconnected, open thermodynamic network of component parts which performs work, existing in a complex regime which combines stability and adaptability in the phase transition between order and chaos, as a plant, animal, fungus, or microbe". By term comparison and removal of terrestrial notions such as plant, animal, fungus, or microbe, Macklem and Seely's definition of life can be reduced to a "compartmentalized, regenerative, metabolic system with secluded in-teractions that is adaptable". Therefore, Macklem and Seely's definition is also a proper subset of Koshland's definition.

Patrick Forterre [24] focuses on viruses and argues that virus is a form of life even though viruses do not possess native self-replication capabilities. Instead, Forterre argues that a virus is essentially an assembly of organs and when the virus infects a host cell, the virus converts the cell into a viral organism. Hence, Forterre defines life as "the mode of existence of living organisms" or "the model of existence of ribosomal and capsid encoding organisms" where "a living organism can be defined as a collection of integrated organs (molecular machines/structures) producing individuals evolving through natural selection". In order to be "model of existence", it implies that there is a program or a blueprint to convert a living entity for its own existence. In the case of a virus, the virus contains a program (a viral genome) can that convert the infected cell into a viral organism for the reproduction of viruses. Under this definition, we can see that living organisms are individually compartmentalized and secluded entities, or organs that can come together for the purpose of improvisation.

In addition, we will argue that Forterre's definition is complementary to Koshland's definition. The seven pillars of life by Koshland [16] acts to clarify the mode or model of existence in Forterre's definition by providing a list of characteristics or properties on this model. Besides affirming that virus are considered alive, it also seems that Forterre [24] had expanded on both Koshland's definition and Carroll's "simplest life form" to include constituent components leading up to the a complete living organism, as long as these constituent components are able to be assembled and produce a living organism. This extension is only required when said individual constituent components do not fulfill the criteria of Koshland's definition. As a result, Forterre's definition has two parts – the full model and the constituent sub-models, which can be integrated to give a full model. It can be inferred from Forterre's definition that the full model forms the core of his definition. If the full model is not a living system, any sub-models derived from this model will logically unable to be assembled into model that is living. However, the inverse may not be true. Even if the full model is living, if the sub-models are not able to be re-assembled into a fully functional model or does not have a program, then the sub-models will not be considered alive. Hence, there implies an emphasis of re-assembly into a living whole. This concept of sub-modeling is recursive. There can always be sub-models from a living sub-model, until it hits the level of sub-atomic particles. This implies that models at the lowest level that constitutes life will fail the full model criteria of Koshland's definition but pass the sub-model criteria of Forterre's definition. Any sub-models from the lowest level of life that pass Forterre's definition but fail Koshland's definition will not be considered alive. In addition, there may also be cases whereby all sub-models from full model fail Forterre's definition, suggesting all sub-models are not considered alive even though the next higher hierarchy passes both Koshland's definition and Forterre's definition.

Hence, we propose a new definition of life based on Koshland [16] and Forterre [24] – *Any system with a boundary to confine the system within a definite volume and protect the system from external effects, consisting of a program that is capable of improvisation, able to react and adapt to the environment, able to regenerate parts of itself or its entirety, with energy system comprises of non-interference sets of secluded reactions for self-sustenance, is considered alive or a living system. Any incomplete system containing a program and can be re-assembled into a living system; thereby, converting the re-assembled system for the purpose of the incomplete system, are also considered alive.*

We do not see our definition as an endpoint but a modest contribution to the discussion. However, it may be possible to test a definition if we consider 2 fundamental effects of life – continuity of the overall system and potential discontinuity of elements within the system. Life as a whole has to continue in one form or another. However, individual elements making up the system may be discontinued or die. This suggests that any definition of life must result in continuity.

## 3 More Alive than Considered

In this section, we shall evaluate some common phenomena and systems as to whether they can be considered alive using the above definition as a checklist. This evaluation neither aim to be exhaustive or comprehensive but provides a starting ground for further debate as to whether the above definition is too broad such that everything is alive or too narrow where intuitively accepted living systems are rejected. The former and latter cases can be known as false positives and false negatives respectively. In addition, the focus of this communica-tion is on the question – Can digital organisms be considered as living? If so, are digital organisms at the boundaries of life? Hence, we will steer the arguments towards answering these questions.

We begin with an obvious example – is *Escherichia coli*, a common intestinal bacterium, or any bacterium alive? *Escherichia coli* has a definite boundary cell membrane and cell wall. Being a bacterium, it has its own genome, which had been sequenced. It had been demonstrated experimentally to adapt to chemicals such as salt [25, 26], food additives [27] and antibiotics [28]. Such adaptations had resulted in changes within the genome [27]. It is able to grow and divide; thus, fulfilling the regeneration criteria. It primarily uses glucose, if available, as its energy source [29]. Even though there is no distinct physical seclusion of chemical reactions within the cell, there are protein scaffolds to isolate chemical reactions [30]. Therefore, *Escherichia coli* fulfills the full model; thus, considered alive.

Are biological viruses alive? As mentioned above, Forterre [24] argued that viruses are alive even though viruses do not fulfill all of Koshland's criteria but as they are able to use the infected cell and convert the cell into a viral organism to suit their own purposes.

Are computer viruses alive? Spafford [31, 32] argued that computer viruses are considered alive by examining each of the criterion proposed by Farmer and Berlin [8]. We shall do the same using the proposed definition in this manuscript. Firstly, a computer virus prior to infecting a file is a list of executable codes in the form of a computer file with a definite beginning and ending. When a computer virus infects a suitable file (host file), the viral code can be linked or integrated into the host file. In this aspect, the behavior of a computer virus is similar to a biological virus undergoing the lysogeny cycle whereby the viral genome is integrated into the genome of the infected host cell, resulting in latent infection. For example, the human immunodeficiency virus (HIV) integrates its genome into human immune cells to form a latent reservoir of virus [33]. In either case, a computer virus has a definite boundary of executable code. Secondly, Spafford [31] argued certain computer viruses can mutate to give variants of the original virus. One mean of computer virus mutation is by employing self-modifying codes, which allows the virus to change its own execution code during execution. Examples of such viruses are Simile and Zmist. Thirdly, a computer virus is separable from the rest of the "non-living" hard disk space. Moreover, computer viruses can be isolated, copied and used to infect another computer. Hence, the fulfillment of the criterion of separation from non-living spaces is evident. Fourthly, Spafford [31] argued that metabolism exists in a computer virus as it uses electrical energy and computational cycles to infect other files. Fifthly, Spafford [31] argued that reproduction is a primary characteristic of computer viruses. We cannot fully agree with Spafford's interpretation. We take the view that a true self-reproducing computer virus must be a virus that infects the bootloader or master boot record. Examples of such computer viruses are Stoned, Brain, Michelangelo, and Elk Cloner. The bootloader, which resides in the master boot record, is the first piece of code to be executed on the computer immediately after BIOS code execution. The purpose of the bootloader is to load the operating system. Hence, when the bootloader is executed, the computer virus can load itself into the memory and activate itself. Conversely, computer viruses that do not infect the bootloader cannot be considered true self-reproducing. Considering that one of the characteristics that separate biological viruses from single-celled organisms is the inability for self-reproduction, we are inclined to consider that bootloader viruses are more capable of self-reproduction than biological virus. Another viewpoint is to consider a computer system as resembling a cell or an organism in the sense that individual components like the power supply or memory chips cannot perform any task on its own. In this case, taking the viewpoint of Forterre [24], all computer viruses have an effect of converting the entire computer system into a computer viral organism. Sixthly, Spafford [31] argued that the computer viruses respond to the environment in the form of altering the flow of execution of infected files examines and changes memory spaces and hard disks for the purpose of infection and spread to other systems. Lastly, computer viruses are lists of executable codes. If we consider each execution to be equivalent to a chemical reaction [34] and a procedure to be an equivalent of a sequence of chemical reactions of execution which is similar to a biochemical pathway, then seclusion of chemical reactions exist by means of procedures.

Therefore, it seems that bootloader-infecting computer viruses can be considered as alive as a bacterium and non-bootloader viruses are at least as alive as biological viruses. If we take the stand that all computer viruses are not able to execute themselves but require the presence of a processor, then we can infer that all computer viruses can convert a computer system into a computer viral organism; thus, implying that all computer viruses are as alive as biological viruses. Given that Forterre [24] considers biological viruses as alive, we do consider computer viruses to be alive.

Spafford [31] concludes his assessment on computer viruses by saying that he will be disappointed if computer viruses are considered the first form of artificial life as their origin is one of unethical practice. In contrast to Spafford [31], we will not be disappointed for two reasons. Firstly, there is growing evidence suggesting that biological viruses pre-dates life [35] due to the observation that virus, as a group, is found to infect all forms of life [36]. In addition, it has been suggested with evidence that DNA, the genomic material or program of all plants, animals and bacteria on earth, may have originated from viruses [37]. In this context, despite having regular bouts of flu (caused by human influenza virus) and an itchy episode of chickenpox (caused by varicella zoster virus), we will not be disappointed to consider that our very existence may be of viral origins. Secondly, early computer viruses, such as Creeper and Elk Cloner, can at worst be considered an annoying practical joke rather than of malicious intent. Bob Thomas, then an engineer at BBN Technologies, to demonstrate the possibility of a mobile program, wrote Creeper. Both Creeper and Elk Cloner display messages on the computer screen but do not corrupt or delete any files or damage the hardware. In today's context, it will almost be equivalent to a friend or relative that keeps repeating the same embarrassing story at every festivity. Hence, despite the scores of damages caused by current computer viruses, its origin is not malicious or truly unethical.

Are digital organisms alive? If computer viruses, which are a form of digital organism or artificial life, are considered alive, is it possible to say that all forms of digital organisms are alive? The link may be to look for a common denominator between biological organisms and digital organisms. From digital organisms, Marion [38] considers computer viruses to be a Turing machine whereby the program modifies its own code during execution and is capable of self-reproduction by making copies of itself. In addition, many scientists [38-44] consider digital organisms or artificial life stimulators to be Turing-complete or Turing-equivalent. From biological organisms, Danchin [45] considers a cell to be a Turing machine and the DNA genome to be equivalent to the computing code or rules for the Turing machine, essentially separating the program from the machine. This is the concept underpinning the work of Lartigue et al. [46] when they synthesized the genome of *Mycoplasma mycoides* chemically and used it to "boot up" a *Mycoplasma capricolum* "casing". Moreover, it is inherent in the properties of biological organisms that self-modification of its program as a result of imprecise copying [47, 48]. Therefore, it seems that any organism, both biological and digital, that can be abstracted as "*a Turing machine that executes a self-modifying code and capable of self-reproduction*" can be considered alive. In this aspect, we come back to a similar minimalistic definition of life proposed by Trifonov [18] – *self-reproduction with variation*. However, this does not imply that all digital organisms are alive.

An interesting example of one of the many other potential forms future digital organisms could take would be a cloud-based or web-based digital entity. Could a cloud-based digital entity be considered alive? That would depend on the nature of said entity. For example, a text file stored on a server and made readily available for cloud distribution could not be considered alive since many, if not none of the criteria are met. It is incapable of self-replication, and any modifications would require external input such as manual editing by a human being. On the other hand, a computer virus stored on a cloud server would be much more alive than said text file. However, since we have already established that a sufficiently sophisticated computer virus could already be considered as alive, it could be suggested that the status of "being stored on a cloud or web-based system and made readily distributable" has no impact on the 'liveliness' of a digital entity as the 'liveliness' of the entity is entirely dependent on its own attributes and capabilities. One could say that the cloud system is merely a route for the digital entity to distribute itself to other connected hardware platforms from the host server much like how an airborne virus (alive) would use aerosols in the air to spread to new hosts from an infected host, while a grain of sand (not alive) is incapable of doing so even though it is exposed to the air just as the virus is; similarly, a computer virus (alive) stored on a cloud server could actively run commands to upload itself from the server to other cloud linked computers without requiring human input, while a simple text file (not alive) would not be able to do so.

One may argue that the existence of the cloud system would mean that a live digital entity existing in such a system is not confined to a fixed boundary, which would render the criterion of having "*a boundary to confine the system within a definite volume*" irrelevant for determining if a particular entity is alive or not. A simple counterargument for this would be the fact that a cloud or web-based system would still require a host server; in fact, all the information on the Internet is stored and hosted on servers for access instead of simply existing in an unknown vacuum or void waiting to magically appear when summoned using web browsers. Hence, the hypothetical alive digital entity is still confined within the boundaries of a host server. Even when said digital entity is distributed to another hardware platform, it is still confined within the new hardware platform. This means that the criterion of having "*a boundary to confine the system within a definite volume*" is still very much relevant when discussing cloud or web-based digital life forms.

Recently, there had been increasing studies on self-awareness, such as developing self-aware robots capable of self-recognition in a mirror [49]. Mowbray and Bronstein [50] defined self-awareness as "observation and computation over the structure, run-time state and external behavior of computing element/system" and Berns and Ghosh [51] considered that self-awareness is a property with no impact on the behavior of a system. For example, in the case of the robots in Gorbenko et al. [49], self-awareness is not required for self-recognition as long as the robots are aware of its surrounding and able to identify the reflective property of a mirror – robot moves, image moves accordingly. Hence, the nature and impact of self-awareness is a matter of debate. Yet, being a physical machine, such a robot has a definite boundary enclosing its electrical and mechanical systems. Gorbenko et al. [49] had demonstrated learning and adaptive abilities in their study. However, there is contentious argument as to how human-like such robots can be [52]. Despite mechanical self-repair and replication had not been demonstrated, it is highly conceivable that given a junkyard of parts, the robot may be able to build a replicate of itself if it was built for the purpose of doing so. Hence, it is likely that this may be at the boundary of life.

Based on Mowbray and Bronstein [50]'s definition of self-awareness, an artificial immune system [53] can be considered to be self-aware as it can observe a compute against its run-time states and external changes by increasing a specific pool of immune "cells". This process is known as clonal selection. Hence, an artificial immune system is a self-aware system by Mowbray and Bronstein [50]'s definition and with all the properties of life as digital organisms; thus, considered alive. Comparing artificial immune system [53], as an example of software-based self-aware system, and self-aware robots [50], as an example of hardware-based self-aware system, one can conclude that it is probably easier to alive on a software platform than on a hardware platform.

On the other hand, is an operating system alive? Perhaps not. Although an operating system can be visualized as a complex sequence of executable codes with a definite boundary, it does not modify itself to be more efficient. Moreover, there is no parallel to say that an operating system can regenerate or reproduce. The only situation close to regeneration is self-recovery from errors and deadlocks. However, we consider this to be over-stretching the interpretation of regeneration. As a result, an operating system fails both the improvisation and regeneration criteria. Thus, an operating system is not considered alive under the full model definition. At the same time, we do not consider an operating system to be alive under the sub-model definition. Although one may argue that an operating system "infects" a group of computer scientists, we consider this to overstretch the interpretation to say that the group of computer scientists is converted into an operating system organism. A susceptible animal cell, such as a bronchial cell, is not able to resist the infection of a target virus, such as an influenza virus. To argue that operating system "infects" a group of computer scientists and convert them into an operating system organism will be saying that the scientists are not able to resist the operating system or unable to change to another field of study or developmental interest. However, this does not mean that an operating system will never be alive. It may be possible to envision a day whereby the operating system is able to improve itself based on usage patterns and possess full error recovery capabilities. Such an operating system will be able to self-optimize its own data structures and codes [54, 55] as a mean of improvisation. At the same time, creating a self-healing computer system is also an area that is being actively researched [56].

Is a laptop alive? A typical modern day laptop would be considered to have fulfilled the criteria of having a boundary which confines the system, which would be the hardware that houses the processors, hard disk drives and other components necessary for a laptop to work; the operating system housed within said components would be the program of the laptop, analogous to the "spark of life" or "soul" concept in complex biological organisms. However, a typical laptop is (at the moment, and likely would remain so in the near future) still incapable of self-improvisation, nor can it react or adapt to external stimuli; neither can it regenerate nor reproduce itself. The operating system contained within is also still incapable of such feats at the moment, as discussed above. Other than that, as an operating system is the core to a laptop, the logical conclusion would be "a laptop cannot be considered alive unless the operating system housed within is first judged to be alive". Since the general consensus is that current operating systems have not yet reached the level where they can be considered alive, current laptops cannot be considered alive. Perhaps in the future, operating systems will become capable of directing machinery to assemble new laptop "bodies" and uploading, downloading or copying themselves into these "hardware platforms". It is probably safe to say that by then, a laptop could be considered to be truly alive.

Are scientific papers alive? Scientific papers are similar to Usenet postings [57] in a different ecology. A scientific paper has a boundary and the scientific content can be considered as its program, which can be considered to improvise as one scientific paper builds on the content of previous papers. Hence, reproduction is present if we consider the reference links as successful reproductive links. If a paper is never cited, we can consider that paper to have failed in its reproduction efforts. In this case, we can even measure reproductive success by citation counts. A scientific paper has different sections, which can be considered as seclusion of processes. However, a scientific paper fails in the area of metabolism and responses to the environment. Similarly, it will be over-stretching interpretation to consider that a scientific paper "infects" scientists and convert them into "paper organisms". Therefore, neither Usenet postings [57] nor scientific papers can be considered alive even though the former had been studied as a form of digital organism. This presents an example of a form digital organism not considered to be alive.

Is a song alive? A song can be seen as an entity made up of human ideas, similar to a scientific paper and Usenet postings. Each song has a boundary in the form of the tune, while the lyrics could be considered the program, which can be modified by changing words in the lyrics, or even writing up completely new lyrics to fit the tune, which is the boundary. The popularity of a song powers it, while the acts of singing, distributing or broadcasting the song could arguably be seen as a form of reproduction if the act teaches other humans said song. It is even possible to argue that a song can regenerate. For example, a song long forgotten with only snatches and phrases remaining in human memory may be recalled once more, or rediscovered from records both written and recorded. However, the song itself has no control of these events, and is also unable to react and adapt by itself. Thus, going by the proposed definition, it would be rather far-fetched to claim that a song is alive. However, it is interesting to note that it does indeed fulfill several criteria of the proposed definition, albeit in its own highly arguable manner and terms.

## 4 Concluding Remarks

The definitions of life by Koshland [16] and Forterre [24] appear to be complementary. These two definitions are merged into a more complete definition. Any entity that fulfills all of the seven criteria below is considered alive: (1) There must be a program, which is an organized plan that describes both the ingredients themselves and the kinetics of the interactions among ingredients as the living system persists through time. (2) There must be a way to change or improvise this program. (3) The living entity must be separable from the non-living spaces. (4) There must be a source of energy to enable some form of metabolism. (5) The entity must be able to regenerate or reproduce to compensate for the wear and tear on a living system. (6) The entity must be able to adapt to environmental hazards. (7) Each set of chemical process must be secluded from each other to prevent interference. Other than that, an entity which does not fulfill all of the seven criteria above, but contains a program, can be re-assembled into a living system, and convert the re-assembled system for the purpose of the incomplete program containing entity is also considered to be alive.

We find that bootloader computer viruses fulfill all seven criteria of Koshland's definition while non-bootloader viruses are considered alive under the sub-model definition. By comparing bacteria to a computer system, another definition of life can be, "*a Turing machine that executes a self-modifying code and capable of self-reproduction*", which can be fulfilled by any computer virus. This definition is similar to the minimalistic defini-tion proposed by Trifonov [18]. Digital organisms that can be abstracted as "a Turing machine that executes a self-modifying code and capable of self-reproduction" may; thus, be considered to be as alive as a bacterium. This in turn suggests that a computer virus as a digital organism can be considered as alive as a bacterium or as alive as a biological virus, depending on the definition; hence, implies that digital organisms should not be considered to be at the boundary of life.

## Appendix A: Definitions of Life

The following is a collated list of various definitions of life. Definition numbers 1 to 95 were extracted from Popa (2003). Definition numbers 96 to 128 were additional definitions extracted from Barbieri (2002), which were not found in Popa (2003). Definition numbers 129 to 135 are some recent (post-2002) definitions of life.

| No. | Year | Proposer | Definition |
|-----|------|----------|------------|
| 1 | 1855 | Virchov | Life will always remain something apart, even if we should find out that it is mechanically aroused and propagated down to the minute detail. |
| 2 | 1871 | Beale | Life is a power, force, or property of a special and peculiar kind, temporarily influencing matter and its ordinary force, but entirely different from, and in no way correlated with, any of these. |
| 3 | 1872 | Bastian | Living things are peculiar aggregates of ordinary matter and of ordinary force which in their separate states do not possess the aggregates of qualities known as life. |
| 4 | 1878 | Bernard | Life is neither a principle nor a resultant. It is not a principle because this principle, in some way dormant or expectant, would be incapable of acting by itself. Life is not a resultant either, because the physicochemical conditions that govern its manifestation cannot give it any direction or any definite form. [. . .] None of these two factors, neither the directing principle of the phenomena nor the ensemble of the material conditions for its manifestation, can alone explain life. Their union is necessary. In consequence, life is to us a conflict. |
| 5 | 1878 | Bernard | If I had to define life in a single phrase ... I should say: Life is creation. |
| 6 | 1880 | Engels | No physiology is held to be scientific if it does not consider death an essential factor of life. [. . .] Life means dying. |
| 7 | 1884 | Spencer | The broadest and most complete definition of life will be the continuous adjustment of internal relations to external relations. |
| 8 | 1923 | Putter | It is the particular manner of composition of the materials and processes, their spatial and temporal organization which constitute what we call life. |
| 9 | 1933 | Von Bertalanffy | A living organism is a system organized in hierarchical order [. . .] of a great number of different parts, in which a great number of processes are so disposed that by means of their mutual relations with wide limits, with constant change of the materials and energies constituting the system and also in spite of disturbances conditioned by external influences, the system is generated or remains in the state characteristic of it, or these processes lead to the production of similar systems. |
| 10 | 1934 | Webster's International Dictionary | Life has the following characteristics: 1. Character of animal or plant manifested by the metabolism, growth, reproduction, and internal powers of adaptation to the environment; 2. vital force distinguished from inorganic matter; 3. experience of animal from birth to death; 4. conscious existence; 5. of being alive; 6. duration of life; 7. individual experience; 8. manner of living; 9. life of the company; 10. the spirit and 11. a duration of similarity. |
| 11 | 1944 | Schrodinger | Life is replication plus metabolism. Replication is explained by the quantum-mechanical stability of molecular structures, while metabolism is explain by the ability of a living cell to extract negative entropy from its surroundings in accordance with the laws of thermodynamics. |
| 12 | 1948 | Alexander | The essential criteria of life are twofold: (1) the ability to direct chemical change by catalysis; (2) the ability to reproduce by autocatalysis. The ability to undergo heritable catalysis changes is general, and is essential where there is competition between dif- |

| No. | Year | Proposer | Definition |
|---|---|---|---|
| | | | ferent types of living things, as has been the case in the evolution of plants and animals. |
| 13 | 1948 | Von Neumann | Life is not one thing but two, metabolism and replication, [. . .] that are logically separable. |
| 14 | 1952 | Perrett | Life is a potentially self-perpetuating open system of linked organic reactions, catalyzed stepwise and almost isothermally by complex and specific organic catalysts which are themselves produced by the system. |
| 15 | 1956 | Hotchkiss | Life is the repetitive production of ordered heterogeneity. |
| 16 | 1959 | Horowitz | The three properties mutability, self-duplication and heterocatalysis comprise a necessary and sufficient definition of living matter. |
| 17 | 1961 | Oparin | Any system capable of replication and mutation is alive. |
| 18 | 1967 | Bernal | Life is a partial, continuous, progressive, multiform and conditionally interactive, self-realization of the potentialities of the atomic electron state. |
| 19 | 1968 | Von Bertalanffy | Life is a hierarchical organization of open systems. |
| 20 | 1972 | Gatlin | Life is a structural hierarchy of functioning units that has acquired through evolution the ability to store and process the information necessary for its own reproduction. |
| 21 | 1973 | Fong | Life is made up of three basic elements: matter, energy and information. [. . .] Any element in life that is not matter and energy can be reduced to information. |
| 22 | 1973 | Maturana and Varela | Life is a metabolic network within a boundary. All that is living must be based on autopoiesis, and if a system is discovered to be autopoietic, that system is defined as living, i.e., it must correspond to the definition of minimal life. |
| 23 | 1973 | Orgel | Living organisms are distinguished by their specified complexity. |
| 24 | 1974 | Ganti | The criteria of living systems are: metabolism, self-reproduction and spatial proliferation. The more complicated kinds also have the ability to mutate and evolve. |
| 25 | 1975 | Maynard-Smith | We regard as alive any population of entities which has the properties of multiplication, heredity and variation. |
| 26 | 1979 | Webster | Life is the property of plants and animals which makes it possible for them to take in food, get energy from it, grow, adapt themselves to their surroundings and reproduce their kind. It is the quality that distinguishes an animal or plant from inorganic matter. Life is the state of possessing this property. |
| 27 | 1979 | Folsome | Life is that property of matter that results in the coupled cycling of bioelements in aqueous solution, ultimately driven by radiant energy to attain maximum complexity. |
| 28 | 1980 | Prigogine | Living units are viewed as objects built up of organic compounds, as dissipative structures, or at least dynamic low entropy systems significantly displaced from thermodynamic equilibrium. |
| 29 | 1981 | Mercer | The sole distinguishing feature, and therefore the defining characteristic, of a living organism is that it is the transient material support of an organization with the property of survival. |
| 30 | 1982 | Haukioja | A living organism is defined as an open system which is able to fulfill the following condition: it is able to maintain itself as an automaton. [. . .] The long-term functioning of automata is possible only if there exists an organization building new automata. |
| 31 | 1984 | Schuster | The uniqueness of life seemingly cannot be traced down to a single feature which is missing in the non-living world. It is the simultaneous presence of all the characteristic properties [. . .] and eventually many more, that makes the essence of a biological system. |
| 32 | 1985 | Csanyi and Kampis | Replication – a copying process achieved by a special network of interrelatedness of components and component-producing processes that produces the same network as that which produces them – characterizes the living organism. |
| 33 | 1986 | Horowitz | Life is synonymous with the possession of genetic properties. Any system with the capacity to mutate freely and to reproduce its mutation must almost inevitably evolve in directions that will ensure its preservation. Given sufficient time, the system will acquire the complexity, variety and purposefulness that we recognize as being alive. |

| No. | Year | Proposer | Definition |
|---|---|---|---|
| 34 | 1986 | Katz | Life is characterized by maximally-complex determinate patterns, patterns requiring maximal determinacy for their assembly. [. . .] Biological templates are determinant templates, and the uniquely biological templates have stability, coherence, and permanence. [. . .] Stable template reproducibility was the great leap, for life is matter that learned to recreate faithfully what are in all other respects random patterns. |
| 35 | 1986 | Sattler | A living system is an open system that is self-replicating, self-regulating, and feeds on energy from the environment. |
| 36 | 1987 | Lifson | Just as wave–particle duality signifies microscopic systems, irreversibility and trend toward equilibrium are characteristic of thermodynamic systems, space-symmetry groups are typical for crystals, so do organization and telemony signify animate matter. Animate, and only animate matter can be said to be organized, meaning that it is a system made of elements, each one having a function to fulfill as a necessary contribution to the functioning of the system as a whole. |
| 37 | 1989 | Vilee et al. | The characteristics that distinguish most living things from nonliving things include a precise kind of organization, a variety of chemical reactions we term metabolism, the ability to maintain an appropriate internal environment even when the external environment changes (a process referred to as homeostasis), movement, responsiveness, growth, reproduction and adaptation to environmental change. |
| 38 | 1992 | Starr and Taggart | To biologists, life is an outcome of ancient events that led to the assembly of nonliving materials into the first organized, living cells. 'Life' is a way of capturing and using energy and materials. 'Life' is a way of seeing and responding to specific changes in the environment. 'Life' is a capacity to reproduce; it is a capacity to follow programs of growth and development. And 'life' evolves, meaning that details in the body plan and functions of each kind of organism can change through successive generations. |
| 39 | 1993 | de Loof | Life is the ability to communicate. |
| 40 | 1993 | Kauffman | Life is an expected, collectively self-organized property of catalytic polymers. |
| 41 | 1994 | NASA working definition of life | Life is a self-sustained chemical system capable of undergoing Darwinian Evolution. |
| 42 | 1994 | Lazcano | Life is like music; you can describe it but not define it. |
| 43 | 1997 | Baltscheffsky | Life may [. . .] be described as a flow of energy, matter and information. |
| 44 | 1997 | Jibu et al. | It is suggested that the existence of the dynamically ordered region of water realizing a boson condensation of evanescent photons inside and outside the cell can be regarded as the definition of life. |
| 45 | 1997 | Root-Bernstein and Dillon | Living organisms are systems characterized by being highly integrated through the process of organization driven by molecular (and higher levels of) complementarity. |
| 46 | 1998 | Clark and Kok | First we give the definition of a biosystem as an adaptive, complex, dynamic system that is alive to some degree. |
| 47 | 1998 | Luisi | A living entity is defined as a system which, owing to its internal process of component production and coupled to the medium via adaptative changes, persists during the time history of the system. |
| 48 | 1999 | Bergareche and Ruiz-Mirazo | Life is seen as a recursive (self-producing and self-reproducing) organization where dynamic and informational levels are mutually dependent. |
| 49 | 1999 | Turian | To Schrodinger's mother of all questions 'What is life', biologists can therefore answer today that they do not consider it some magical force that animated lifeless materials, but rather an emergent property based on the behavior of the materials that make up living things. |
| 50 | 2000 | Dyson | Life is defined as a material system that can acquire store, process, and use information to organize its activities. |
| 51 | 2000 | Kunin | Life is defined as a system of nucleic acid and protein polymerases with a constant supply of monomers, energy and protection. |
| 52 | 2001 | Hazen | A potentially useful conceptual approach to the question of life's definition is to consider the origin of life as a sequence of 'emergent' events, each of which adds to molecular complexity and order. |

| No. | Year | Proposer | Definition |
| --- | --- | --- | --- |
| 53 | 2001 | Korzeniewski | Life on the Earth [. . .] seems to possess three properties (strongly related to each other and in fact being different aspects of the same thing) which are absent in inanimate systems. Namely, life is (1) composed of particular individuals, that (2) reproduce (which involves transferring their identity to progeny) and (3) evolve (their identity can change from generation to generation). A living individual is defined as a network of inferior negative feedbacks (regulatory mechanisms) subordinated to (being at the service of) a superior positive feedback (potential of expansion of life). |
| 54 | 2001 | Sertorio and Tinetti | We adopt this weak definition of life. A living system occupies a finite domain, has structure, performs according to an unknown purpose, and reproduces itself. |
| 55 | 2001 | Yang et al. | The characteristics of artificial life are emergence and dynamic interaction with the environment. |
| 56 | 2002 | Abel | Ignoring the misgivings of those few life-origin theorists with 'mule' fixations, life is the 'symphony' of dynamic and highly integrated algorithmic processes which yields homeostatic metabolism, development, growth, and reproduction. |
| 57 | 2002 | Altstein | Life is the process of existence of open non-equilibrium complete systems, which are composed of carbon-based polymers and are able to self-reproduce and evolve on the basis of template synthesis of their polymer components. |
| 58 | 2002 | Anbar | Any living system must comprise four distinct functions: 1. Increase of complexity; 2. directing the trends of increased complexity; 3. preserving complexity; and 4. recruiting and extracting the free energy needed to drive the three preceding motions. |
| 59 | 2002 | Arrhenius | Life is defined as a system capable of 1. self-organization; 2. Self-replication; 3. evolution through mutation; 4. metabolism and 5. concentrative encapsulation. |
| 60 | 2002 | Baltcheffsky | Life is defined as a self-sustained molecular system transforming energy and matter, thus realizing its capacity of replication with mutations and anastrophic evolution. |
| 61 | 2002 | Boiteau | Life appears as a set of symbiotically-linked molecular engines, permanently operating out of equilibrium, in an open flow of energy and matter, although recycling a great deal of their own chemical components, through cyclic chemistry. |
| 62 | 2002 | Brack | Life is a chemical system capable of transferring its molecular information independently (self-reproduction) and also capable of making some accidental errors to allow the system to evolve (evolution). |
| 63 | 2002 | Buick | In order to be recognizable life must: 1. be a non-equilibrium chemical system; 2. contain organic polymers; 3. reproduce itself; 4. metabolize by itself; 5. be segregated from the environment. |
| 64 | 2002 | Eirich | We consider to be alive any homo- or heterotrophic cellular irreversible heat engine, or their assembly, that carries instructions for its function, reproduction, topical location, individuality and life cycle. |
| 65 | 2002 | Erokhin | The living organism is a multilevel open catalytic system achieved in its evolutionary development of maximal catalytic activity in basic process and possessing the property of self-reproduction. Life is a process of functioning of living organisms. |
| 66 | 2002 | Friedman | Paraphrasing Theodosius Dobzhansky: Life is what the scientific establishment (probably after some healthy disagreement) will accept as life. |
| 67 | 2002 | Guerrero | Life is matter that makes choices, binds time and breaks gradients. |
| 68 | 2002 | Guimaraes | Living beings are complex functional systems. Life is an abstract concept describing properties of cells, concrete objects. Life is the process manifested by individualized evolutionary metabolic systems. The functions, which are called life, are: metabolism, growth, and reproduction with stability through generations. |
| 69 | 2002 | Gusev | Life is an energy-dependent chemical cyclic process which results in an increase of functional and structural complexity of living systems and their inhabited environment. |
| 70 | 2002 | Hennet | Life is simply a particular state of organized instability. |
| 71 | 2002 | Horowitz | Life is synonymous with the possession of genetic properties, i.e., the capacities for self-replication and mutation. |

| No. | Year | Proposer | Definition |
|---|---|---|---|
| 72 | 2002 | Kawamura | Life is a system which has subjectivity. |
| 73 | 2002 | Keszthelyi | Life is metabolism and proliferation. |
| 74 | 2002 | Klabunovsky | Life is an inevitable consequence of the existence of optically active organic compounds (like proteins). |
| 75 | 2002 | Kolb | Life is a new quality brought upon an organic chemical system by a dialectic change resulting from an increase in the quantity of complexity of the system. This new quality is characterized by the ability of temporal self-maintenance and self-preservation. |
| 76 | 2002 | Kompanichenko | Life is a highly organized form of intensified resistance to spontaneous processes of destruction developing by means of expedient interaction with the environment and regular self-renovation. |
| 77 | 2002 | Krumbein | Any system that creates, maintains and/or modifies dissymmetry is Alive. |
| 78 | 2002 | Kulaev | Life is the form of existence of a substance capable of self-reproduction and maintenance of permanent metabolism with the environment. |
| 79 | 2002 | Lacey et al. | To be alive is to be a degradable, semi-isolated system which has survived because it was able to generate a molecular memory both of its environment and how to respond to it. Life can be defined by the following list of characteristics: 1. It must be isolated from an external environment but still be able to exchange materials with it (must possess individuality). 2. It must be susceptible to degradation by the environment. 3. It must be small enough so that rates of transfer of materials into and out of the isolated system will be rapid. 4. It must be able to generate a molecular representation of its internal and external environments (i.e., it must have a molecular memory of its environments). 5. It must be able to sense the environment and respond to it, i.e., it must be able to synthesize active molecules capable of utilizing materials it encounters in the environment. |
| 80 | 2002 | Lahav | A terrestrial living entity is an ensemble of molecular-informational feedback-loop systems consisting of a plurality of organic molecules of various kinds, coupled spatially and functionally by means of template-and-sequence directed networks of catalyzed reactions and utilizing, interactively, energy and inorganic and organic molecules from the environment. A living entity is an uninterrupted succession of ensembles of feedback-loop systems evolved between the emergence time and the moment of observation. |
| 81 | 2002 | Lahav and Nir | A living entity is an ensemble of molecules which exhibit spatial organization and molecular-informational feedback loops in utilization of materials and energy from the environment for its growth, reproduction and evolution. |
| 82 | 2002 | Lauterbur | It's alive if it can die. |
| 83 | 2002 | Marko | From a chemical point of view, life is a complex autocatalytic process. This means that the end products of the chemical reactions in a living cell (nucleic acids, polypeptides and proteins, oligo- and polysaccharides) catalyze their own formation. From a thermodynamical point of view, life is a mechanism which uses complex processes to decrease entropy. |
| 84 | 2002 | Nair | Life is an attribute of living systems. It is continuous assimilation, transformation and rearrangement of molecules as per an in-built program in the living system so as to perpetuate the system. |
| 85 | 2002 | Nealso | Any definition of life that is useful must be measurable. We must define life in terms that can be turned into measurables, and then turn these into a strategy that can be used to search for life. So what are these? a. structures, b. chemistry, c. replication with fidelity and d. evolution. |
| 86 | 2002 | Noda | Life is a system which can reproduce itself using genetic mechanisms. |
| 87 | 2002 | Polishchuck | Life is a structurally stable negentropy current supported by self-correction for the biological hereditary genetic code [. . .] providing an energy inflow. |
| 88 | 2002 | Rizzotti | Life is instantiated by the objects that resist decay by means of constructive assimilation. |
| 89 | 2002 | Schulze-Makuch et al. | We propose to define living systems as those that are: (1) composed of bounded mi- |

| No. | Year | Proposer | Definition |
|-----|------|----------|------------|
|     |      |          | cro-environments in thermodynamic equilibrium with their surroundings; (2) capable of transforming energy to maintain their low-entropy states; and (3) able to replicate structurally distinct copies of themselves from an instructional code perpetuated indefinitely through time despite the demise of the individual carrier through which it is transmitted. |
| 90  | 2002 | Scorei | Life is a form of matter organization that is energetically and informationally self-supported, with a good capacity of self-instruction and creation. |
| 91  | 2002 | Soriano | Life is the ability of an organism to formulate questions. |
| 92  | 2002 | Valenzuela | Life is a historical process of anagenetic organizational relays. |
| 93  | 2002 | Von Kiedrowski | Life is a population of functionally connected, local, non-linear, informationally-controlled chemical systems that are able to self-reproduce, to adapt, and to coevolve to higher levels of global functional complexity. |
| 94  | 2002 | Wong | A living system is one capable of reproduction and evolution, with a fundamental logic that demands an incessant search for performance with respect to its building blocks and arrangement of these building blocks. The search will end only when perfection or near perfection is reached. Without this built-in search, living systems could not have achieved the level of complexity and excellence to deserve the designation of life. |
| 95  | 2002 | Yockey | The existence of a genome and the genetic code divides living organisms from non-living matter. |
| 96  | 1802 | Lamarck | Life is an order or a state of things in the component parts of a body that makes organic movement possible and that effectively succeeds, as long as it persists, in opposing death. |
| 97  | 1855 | Buchner | Spontaneous generation exists, and higher forms have gradually and slowly developed from previously existing lower forms, always determined by the state of the earth, but without immediate influence of a higher power. |
| 98  | 1866 | Haechel | Any detailed hypothesis concerning the origin of life must, as yet, be considered worthless, because up till now we have no satisfactory information concerning the extremely peculiar conditions which prevailed on the earth at the time when the first organisms developed. |
| 99  | 1868 | Huxley | The vital forces are molecular forces. |
| 100 | 1868 | Liebig | We may only assume that life is just as old and just as eternal as matter itself … Why should not organic life be thought of as present from the very beginning just as carbon and its compounds, or as the whole of uncreatable and indestructible matter in general? |
| 101 | 1869 | Browning | There is no boundary line between organic and inorganic substances… Reasoning by analogy, I believe that we shall before long find it an equally difficult task to draw a distinction between the lowest forms of living matter and dead matter. |
| 102 | 1890 | Weismann | The living organism has already been compared with a crystal, and the comparison is, mutatis mutandis, justifiable. |
| 103 | 1987 | Pfeffer | Even the best chemical knowledge of the bodies occurring in the protoplasm no more suffices for the explanation and understanding of the vital processes, than the most complete chemical knowledge of coal and iron suffices for the understanding of a steam engine. |
| 104 | 1908 | Macallum | When we seek to explain the origin of life, we do not require to postulate a highly complex organism … as being the primal parent of all, but rather one which consists of a few molecules only and of such a size that it is beyond the limit of vision with the highest powers of the microscope. |
| 105 | 1924 | Oparin | What are the characteristics of life? In the first place there is a definite structure or organisation. Then there is the ability of organisms to metabolise, to reproduce others like themselves, and also their response to stimulation. |
| 106 | 1929 | Woodger | It does not seem necessary to stop at the word "life" because this term can be eliminated from the scientific vocabulary since it is an indefinable abstraction and we can |

| No. | Year | Proposer | Definition |
|-----|------|----------|------------|
|  |  |  | get along perfectly well with "living organism" which is an entity which can be speculatively demonstrated. |
| 107 | 1933 | Bohr | The existence of life must be considered as an elementary fact that cannot be explained, but must be taken as a starting point in biology, in a similar way as the quantum of action, which appears as an irrational element from the point of view of classical physics, taken together with the existence of elementary particles, form the foundation of atomic physics. |
| 108 | 1966 | Muller | It is alive any entity that has the properties of multiplication, variation and heredity. |
| 109 | 1970 | Monod | Living beings are teleonomic machines, self-constructing machines and self-reproducing machines. There are, in other words, three fundamental characteristics common to all living beings: teleonomy, autonomous morphogenesis and invariant reproduction. |
| 110 | 1973 | Orgel | Living beings are CITROENS (Complex Information-Transforming Reproducing Objects that Evolve by Natural Selection). |
| 111 | 1977 | Argyle | Life on earth today is a highly degenerate process in that there are millions of different gene strings (species) that spell the one word "life". |
| 112 | 1981 | Eigen | The most conspicuous attribute of biological organization is its complexity… The problem of the origin of life can be reduced to the question: "Is there a mechanism of which complexity can be generated in a regular, reproducible way?" |
| 113 | 1988 | Edelman | Animate objects are self-replicating systems containing a genetic code that undergoes mutation and whose variant individuals undergo natural selection. |
| 114 | 1989 | Langton | Artificial Life can contribute to theoretical biology by locating life as-we-know-it within the larger picture of life-as-it-could-be. |
| 115 | 1992 | Belin and Farmer | Life involves: (1) a pattern in spacetime (rather than a specific material object); (2) self-reproduction, in itself or in a related organism; (3) information-storage of a self-representation; (4) metabolism that converts matter/energy; (5) functional interactions with the environment; (6) interdependence of parts within the organism; (7) stability under perturbations of the environment; and (8) the ability to evolve. |
| 116 | 1994 | Emmeche | Life itself is a computational phenomenon. |
| 117 | 1996 | Brack | Life is a chemical system capable to replicate itself by autocatalysis and to make errors which gradually increase the efficiency of autocatalysis. |
| 118 | 1996 | Fox | Life consists of proteinaceous bodies formed of one or more cells containing membranes that permit it to communicate with its environment via transfer of information by electrical impulse or chemical substance, and is capable of morphological evolution by self-organisation of precursors, and displays attributes of metabolism, growth, and reproduction. This definition embraces both protolife and modern life. |
| 119 | 1996 | Ganti | At the cellular level the living systems are proliferating, program-controlled fluid chemical automatons, the fluid organisation of which are chemoton organisation. And life itself – at the cellular level – is nothing else but the operation of these systems. |
| 120 | 1996 | Hoffmeyer | The basic unit of life is the sign, not the molecule. |
| 121 | 1996 | Igamberdiev | Life is a self-organised and self-generating activity of open non-equilibrium systems determined by their internal semiotic structure. |
| 122 | 1996 | Varela | A physical system can be said to be living if it is able to transform external energy/matter into an internal process of self-maintenance and self-generation. This common sense, macroscopic definition, finds its equivalent at the cellular level in the notion of autopoiesis. This can be generalised to describe the general pattern for minimal life, including artificial life. In real life, the autopoietic network of reactions is under the control of nucleic acids and the corresponding proteins. |
| 123 | 1997 | Hucho and Buchner | Signal transduction is as fundamental a feature of life as metabolism and self-replication. |
| 124 | 1998 | Kull | An organism is a text to itself since it requires reading and re-representing its own structures for its existence, e.g. for growth and reparation. It also uses reading of its memory when functioning. This defines an organism as a self-reading text. |

| No. | Year | Proposer | Definition |
|---|---|---|---|
| 125 | 2000 | Yockey | The segregated, linear and digital character of the genetic message is an elementary fact and therefore essentially a definition of life. It is a gulf between living organisms and inanimate matter. |
| 126 | 2001 | Sebeok | Because there can be no semiosis without interpretability – surely life's cardinal propensity – semiosis presupposes the axiomatic identity of the semiosphere with the biosphere. |
| 127 | 2002 | Koshland | If I were in ancient Greece, I would create a goddess of life whom I would call PICERAS, … because there are seven fundamental principles (the Seven Pillars of Life) on which a living system is based: P (Program), I (Improvisation), C (Compartmentalization), E (Energy), R (Regeneration), A (Adaptability), and S (Seclusion). |
| 128 | 2002 | Trifonov | Life is an almost precise replication. |
| 129 | 2004 | Ruiz-Mirazo et al. | 'A living being' is any autonomous system with open-ended evolutionary capacities. |
| 130 | 2006 | Zhuravlev and Avetisov | An explicit definition of life consisting of three parts to finish with the relationship of this definition to the reconstruction of events leading to the emergence of life: (i) Life, as we see it now, is a specific state of matter (the living state) resulting from the interaction between matter and energy carriers. This interaction starts from utilization of solar radiation by autotrophic organisms, and spreads over a diversity of organisms via numerous (bio)chemical cycles. A significant part of the utilized energy is retained in organisms by molecular carriers and "network channels" of high energy content; lessening of the utilized energy pool up to some critical level entails death. (ii) Life on Earth is represented by a specific hierarchical system (the living system) consisting of self-reproducing agents. These agents are the only reference matter of life and are often represented by organisms. They can sometimes be represented as more complex units: bisexual pair, beehive, etc. The agents being individuals can interact each other and therefore the whole system can be considered as the fragmented and the integral entity simultaneously. Different levels of organization of agents correspond to different levels of life hierarchy. Life as a system shows its worth in the diversity of constraints, feedbacks and interconnections with surroundings. (iii) Life on Earth proceeds as the specific process (the living process). It is expressed in transformations of surroundings (by agents) and in transmutations of the self-reproducing agents themselves. From the physico-chemical point of view, the living process has both dynamic and informational contents. It allows the agents to properly meet the changes of environment and to expand (spread) over a space, thus increasing the level of system complexity and differentiation. In our reconstruction of the emergence of life we have to keep the three-sided view of life, as specified above. This reconstruction should start from finding such (molecular?) representatives of the self-reproducing agents, which, being the simplest ones, nevertheless, reserve the competence to create the living state, the living system, and the living process. |
| 131 | 2008 | Carroll | The simplest form of life exists in the form of self-amplifying, autocatalytic reactions. |
| 132 | 2008 | Lazcano | Life can be considered as a self-sustaining chemical system (i.e., one that turns environmental resources into its own building blocks) that is capable of undergoing natural selection. |
| 133 | 2010 | Macklem and Seely | Life as a "self-contained, self-regulating, self-organizing, self-reproducing, interconnected, open thermodynamic network of component parts which performs work, existing in a complex regime which combines stability and adaptability in the phase transition between order and chaos, as a plant, animal, fungus, or microbe". |
| 134 | 2010 | Forterre | Life is the mode of existence of living organisms. Life is the model of existence of ribosomal and capsid encoding organisms. A living organism can be defined as a collection of integrated organs (molecular machines/structures) producing individuals evolving through natural selection. |
| 135 | 2011 | Trifonov | Life is self-reproduction with variations. |

## Author Biography

**Maurice HT Ling** (Contact Author)
School of Chemical and Biomedical Engineering, Nanyang Technological University, Singapore; Department of Zoology, The University of Melbourne, Australia (mauriceling@acm.org)

**Yong Zher Koh**
Department of Biological Sciences, University of Portsmouth, UK.

## References

1. Shaw, J. (2012). Using small group debates to actively engage students in an introductory microbiology course. Journal of Microbiology & Biology Education 13, 2.
2. Jago, R., Zakeri, I., Baranowski, T., & Watson, K. (2007) Decision boundaries and receiver operating characteristic curves: new methods for determining accelerometer cutpoints. Journal of Sports Science 25, 937-44.
3. Lopez-Garcia, P., & Moreira, D. (2012). Viruses in biology. Evolution: Education and Outreach 5, 389-398.
4. Kumar, A. (2012). An overview of anstract and physical characteristics of "artificial life systems". International Journal of Scientific and Research Publications 2, 12.
5. Wolfe-Simon, F., Switzer Blum, J., Kulp, T. R., Gordon, G. W., Hoeft, S. E., Pett-Ridge, J., Stolz, J. F., Webb, S. M., Weber, P. K., Davies, P. C., Anbar, A. D., & Oremland, R. S. (2011) A bacterium that can grow by using arsenic instead of phosphorus. Science 332, 1163-6.
6. Bedau, M. (2012) Introduction to philosophical problems about life. Synthese 185, 1-3.
7. Bedau, M. A., McCaskill, J. S., Packard, N. H., & Rasmussen, S. (2009) Living Technology: Exploiting Life's Principles in Technology. Artificial Life 16, 89-97.
8. Farmer, J. D., & Belin, A. (1992) Artificial life: The coming revolution. In Langton C., Taylor C., Farmer J. and Rasmussen S. (eds) Artificial Life II: Proceedings of the Workshop on Artificial Life.
9. Cimpoiasu V. M., & Popa, R. (2012) Biotic Abstract Dual Automata (BiADA): a novel tool for studying the evolution of prebiotic order (and the origin of life). Astrobiology 12, 1123-34.
10. Deplazes, A., & Huppenbauer, M. (2009) Synthetic organisms and living machines. Systems and Synthetic Biology 3, 55-63.
11. Gayon, J. (2010) Defining Life: Synthesis and Conclusions. Origins of Life and Evolution of Biospheres 40, 231-244.
12. Machery, E. (2012) Why I stopped worrying about the definition of life... and why you should as well. Synthese 185, 145-64.
13. Mayr, J. (1982). The growth of biological thought: Diversity, evolution, and inheritance. Cambridge: The Belknap Press of Harvard University Press.
14. Popa, R. (2003). Between necessity and probability. Searching for the definition and the origin of life. In Advances in Astrobiology and Biogeophysics. Berlin: Springer.
15. Barbieri, M. (2002). The organic codes: an introduction to semantic biology. Cambridge University Press.
16. Koshland, D. E., Jr. (2002). The seven pillars of life. Science 295, 2215-2216.
17. Trifonov, E. N. (2008) Tracing life back to elements. Physics of Life Reviews 5, 121-32.
18. Trifonov, E. N. (2011) Vocabulary of definitions of life suggests a definition. Journal of Biomolecular Structure and Dynamics 29, 259-66.
19. Ruiz-Mirazo, K., Pereto, J., & Moreno, A. (2010) Defining life or bringing biology to life. Origins of Life and Evolution of Biospheres 40, 203-213.
20. Carroll, J. D. (2009) A new definition of life. Chirality 21, 354-8.
21. Lazcano, A. (2008) Towards a definition of life: the impossible quest? Space Science Reviews 135, 5-10.
22. Zhuravlev, Y. N., & Avetisov, V. A. (2006) The definition of life in the context of its origin. Biogeosciences Discussions 3, 155-81.
23. Macklem, P. T., & Seely, A. (2010) Towards a definition of life. Perspectives in Biology and Medicine 53, 330-40.
24. Forterre, P. (2010) Defining life: the virus viewpoint. Origins of Life and Evolution of Biospheres 40, 151-60.
25. Goh, D. J. W, How, J. A., Lim, J. Z. R., Ng, W. C., Oon, J. S. H., Lee, K. C., Lee, C. H., & Ling, M. H. T. (2012). Gradual and step-wise halophilization enables Escherichia coli ATCC 8739 to adapt to 11% NaCl. Electronic Physician 4, 527-535.
26. How, J. A., Lim, J. Z. R., Goh, D. J. W., Ng, W. C., Oon, J. S. H., Lee, K. C., Lee, C. H., & Ling, M. H. T. (2013). Adaptation of Escherichia coli ATCC 8739 to 11% NaCl. Dataset Papers in Biology 2013, Article ID 219095.
27. Lee, C. H., Oon, J. S. H., Lee, K. C., & Ling, M. H. T. (2012). Escherichia coli ATCC 8739 adapts to the presence of sodium chloride, monosodium glutamate, and benzoic acid after extended culture. ISRN Microbiology 2012, Article ID 965356.
28. Capita, R., Alvarez-Fernandez, E., Fernandez-Buelta, E., Manteca, J., & Alonso-Calleja, C. (2013) Decontamination treatments can increase the prevalence of resistance to antibiotics in Escherichia coli naturally present on poultry. Food Microbiology 34, 112-7.
29. Low, S. X. Z., Aw, Z. Q., Loo, B. Z. L., Lee, K. C., Oon, J. S. H., Lee, C. H., & Ling, M. H. T. (2013). Viability of Escherichia coli ATCC 8739 in nutrient broth, luria-bertani broth and brain heart infusion over 11 weeks. Electronic Physician 5, 576-581.
30. Khemici, V., Toesca, I., Poljak, L., Vanzo N. F., & Carpousis, A. J. (2004). The RNase E of Escherichia coli has at least two binding sites for DEAD-box RNA helicases: functional replacement of RhlB by RhlE. Molecular Microbiology 54, 1422-30.
31. Spafford, E. H. (1990) Computer Viruses - A Form of Artificial Life? In Computer Science Technical Reports. Purdue University.
32. Spafford, E. H. (1994) Computer Viruses as Artificial Life. Artificial Life 1, 249-265.
33. Finzi, D., Blankson, J., Siliciano, J. D., Margolick, J. B., Chadwick, K., Pierson, T., Smith, K., Lisziewicz, J., Lori, F., Flexner, C., Quinn, T. C., Chaisson, R. E., Rosenberg, E., Walker, B., Gange, S., Gallant, J., & Siliciano, R. F. (1999). Latent infection of CD4+ T cells provides a mechanism for lifelong persistence of HIV-1, even in patients on effective combination therapy. Nature Medicine 5, 512-517.
34. Yamamoto, L. (2007). Code Regulation in Open Ended Evolution. In Genetic Programming (eds. by Ebner M, O'Neill M, Ekárt A, Vanneschi L & Esparcia-Alcázar A), pp. 271-80. Springer, Berlin, Heidelberg.
35. Holmes, E. C. (2011) What does virus evolution tell us about virus origins? Journal of Virology 85, 5247-51.
36. Forterre, P. (2006) The origin of viruses and their possible roles in major evolutionary transitions. Virus Research 117, 5-16.
37. Pina, M., Bize, A., Forterre, P, & Prangishvili, D. (2011) The archeoviruses. FEMS Microbiological Reviews 35, 1035-1054.

38. Marion, J.Y . (2012). From Turing machines to computer viruses. Philosophical Transactions of The Royal Society, A: Mathematical, Physical and Engineering Sciences 370, 3319-3339.

39. Grabowski, L. M., Bryson, D. M., Dyer, F. C., Pennock, R. T., & Ofria, C. (2011). Clever creatures: case studies of evolved digital organisms. In Advances in Artificial Life, ECAL 2011: Proceedings of the Eleventh European Conference on the Synthesis and Simulation of Living Systems (pp. 276-83).

40. Hutton, T. J. (2010). Codd's Self-Replicating Computer. Artificial Life 16, 99-117.

41. Ling, M. H. (2012a). An artificial life simulation library based on genetic algorithm, 3-character genetic code and biological hierarchy. The Python Papers 7, 5.

42. Ling, M. H. (2012b). Ragaraja 1.0: The Genome Interpreter of Digital Organism Simulation Environment (DOSE). The Python Papers Source Codes 4, 2.

43. Murata, S., & Kurokawa, H. (2012). History of self-organizing machines. In Self-Organizing Robots (pp. 37-57). Springer, Tokyo.

44. Uchida, Y., Sakamoto, M., Taniue, A., Katamune, R., Ito, T., Furutani, H., & Kono, M. (2010). Some properties of four-dimensional parallel Turing machines. Artificial Life and Robotics **15**, 385-388.

45. Danchin, A. (2009). Bacteria as computers making computers. FEMS Microbiological Review **33**, 3-26.

46. Lartigue, C., Vashee, S., Algire, M. A., Chuang, R. Y., Benders, G. A., Ma, L., Noskov, V. N., Denisova, E. A., Gibson, D. G., Assad-Garcia, N., Alperovich, N., Thomas, D. W., Merryman, C., Hutchison, C. A., 3rd, Smith, H. O., Venter, J. C., & Glass, J. I. (2009), Creating bacterial strains from genomes that been cloned and engineered in yeast. Science 325, 1693-1696.

47. Fijalkowska, I. J., Schaaper, R. M., & Jonczyk, P. (2012). DNA replication fidelity in Escherichia coli: a multi-DNA polymerase affair. FEMS Microbiological Reviews 36, 1105-1121.

48. Umar, A., & Kunkel, T. A. (1996). DNA-replication fidelity, mismatch repair and genome instability in cancer cells. European Journal of Biochemistry 238, 297-307.

49. Gorbenko, A., Popov, V., & Sheka, A. (2012) Robot self-awareness: Exploration of internal states. Applied Mathematical Sciences **6**: 675-688.

50. Mowbray M., & Bronstein A. (2005) What kind of self-aware systems does the grid need. HP Laboratories Bristol, HPL-2002-266 (R. 1).

51. Berns, A., & Ghosh, S. Dissecting self-* properties. In Third IEEE International Conference on Self-Adaptive and Self-Organizing Systems, 2009. SASO'09 (pp. 10-19). IEEE, 2009.

52. Wang J. (2012) Will a Robot Be a Human? In Artificial Intelligence and Soft Computing (eds. by Rutkowski L, Korytkowski M, Scherer R, Tadeusiewicz R, Zadeh L & Zurada J), pp. 519-27. Springer Berlin Heidelberg.

53. Hofmeyr S. A., & Forrest S. (2000) Architecture for an artificial immune system. Evolutionary computation **8**, 443-73.

54. Wimmer, C., & Wurthinger, T. (2012). Truffle: a self-optimizing runtime system. In Proceedings of the 3rd annual conference on Systems, programming, and applications: software for humanity (pp. 13-14). ACM, Tucson, Arizona, USA.

55. Wurthinger, T., Wob, A., Stadler, L., Duboscq, G., Simon, D., & Wimmer, C. (2012). Self-optimizing AST interpreters. In Proceedings of the 8th symposium on Dynamic languages (pp. 73-82). ACM, Tucson, Arizona, USA.

56. Niazi, M. A. (2013). Complex adaptive systems modeling: a multidisciplinary roadmap. Complex Adaptive Systems Modeling 1, 1.

57. Best, M. L. (1997). An ecology of text: Using text retrieval to study alife on the Net. Artificial Life 3, 261